\documentclass[letterpaper]{article} % DO NOT CHANGE THIS
\usepackage{aaai24}  % DO NOT CHANGE THIS
\usepackage{times}  % DO NOT CHANGE THIS
\usepackage{helvet}  % DO NOT CHANGE THIS
\usepackage{courier}  % DO NOT CHANGE THIS
\usepackage[hyphens]{url}  % DO NOT CHANGE THIS
\usepackage{graphicx} % DO NOT CHANGE THIS
\usepackage{natbib}  % DO NOT CHANGE THIS AND DO NOT ADD ANY OPTIONS TO IT
\usepackage{caption} % DO NOT CHANGE THIS AND DO NOT ADD ANY OPTIONS TO IT
\usepackage{algorithm}
\usepackage{algorithmic}

\usepackage{newfloat}
\usepackage{listings}
\DeclareCaptionStyle{ruled}{labelfont=normalfont,labelsep=colon,strut=off} % DO NOT CHANGE THIS
\floatstyle{ruled}
\newfloat{listing}{tb}{lst}{}
\floatname{listing}{Listing}
\usepackage{epsfig}
\usepackage{amsmath}
\usepackage{amssymb}
\usepackage{multirow}
\usepackage{booktabs}

\title{Attack Deterministic Conditional Image Generative Models for Diverse and Controllable Generation}
\author{
    Tianyi Chu\textsuperscript{\rm 1},
    Wei Xing\textsuperscript{\rm 1},
    Jiafu Chen\textsuperscript{\rm 1},
    Zhizhong Wang\textsuperscript{\rm 1},
    Jiakai Sun\textsuperscript{\rm 1},
    Lei Zhao\textsuperscript{\rm 1}\thanks{Corresponding author},
    Haibo Chen\textsuperscript{\rm 2},\\
    Huaizhong Lin\textsuperscript{\rm 1}$^*$
}
\affiliations{
    \textsuperscript{\rm 1}Zhejiang University,\\
    \textsuperscript{\rm 2}Nanjing University of Science and Technology\\
	\{chutianyi, wxing, chenjiafu, endywon, csjk, cszhl, linhz\}@zju.edu.cn,\ 
	hbchen@njust.edu.cn
	
}

\usepackage{bibentry}
\begin{document}

\maketitle

\begin{abstract}
	Existing generative adversarial network (GAN) based conditional image generative models typically produce fixed output for the same conditional input, which is unreasonable for highly subjective tasks, such as large-mask image inpainting or style transfer. 
	On the other hand, GAN-based diverse image generative methods require retraining/fine-tuning the network or designing complex noise injection functions, which is computationally expensive, task-specific, or struggle to generate high-quality results.
	\emph{\emph{Given that many deterministic conditional image generative models have been able to produce high-quality yet fixed results, we raise an intriguing question:} is it possible for pre-trained deterministic conditional image generative models to generate diverse results without changing network structures or parameters?} To answer this question, we re-examine the conditional image generation tasks from the perspective of adversarial attack and propose a simple and efficient plug-in projected gradient descent (PGD) like method for diverse and controllable image generation. The key idea is attacking the pre-trained deterministic generative models by adding a micro perturbation to the input condition. In this way, diverse results can be generated without any adjustment of network structures or fine-tuning of the pre-trained models. 
	In addition, we can also control the diverse results to be generated by specifying the attack direction according to a reference text or image. 
	Our work opens the door to applying adversarial attack to low-level vision tasks, and experiments on various conditional image generation tasks demonstrate the effectiveness and superiority of the proposed method.
\end{abstract}
%------------------------------3 diversity ---------------------
\begin{figure*}[t]
	\centering
	\includegraphics[width=0.98\textwidth]{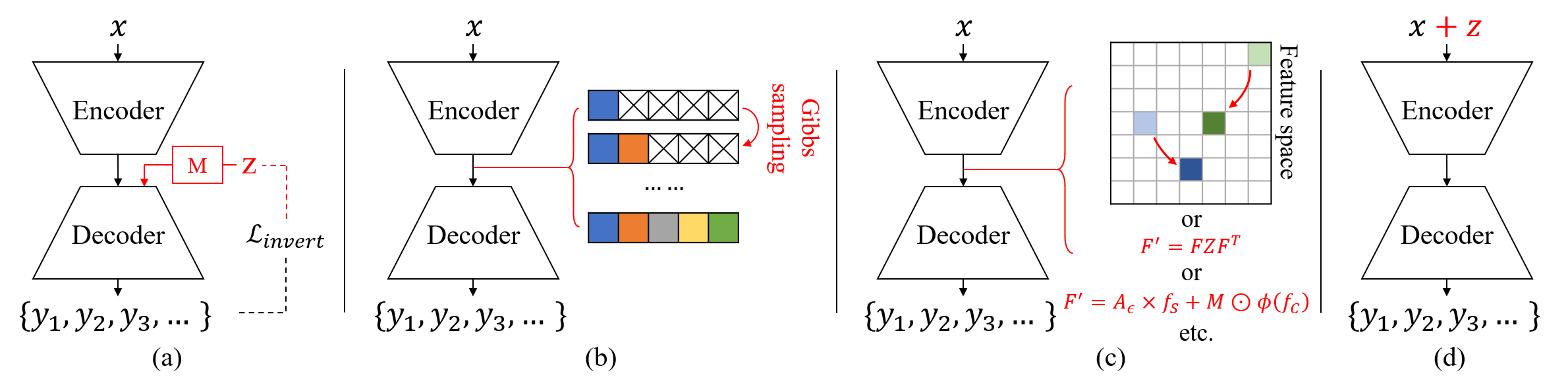}
	\vspace{-3mm}
	\caption{
		Three mainstream methods for introducing diversity in conditional image generation: (a) injecting noise through modulation module, (b) Gibbs sampling on feature sequences, and (c) transforming features according to specific rules. 
		Our proposed method (d) allows pre-trained deterministic generative models to generate diverse results without multiple-step sampling, sophisticated transformation functions, or any adjustments to the network structure or parameters.}
	\label{fig:diff_diversity}
	\vspace{-0mm}
\end{figure*}
%--------------------------------------------------------------
\section{Introduction}
Conditional image generation (e.g., image inpainting, style transfer, super-resolution, denoising) is one of the representative subtasks of image synthesis, which typically uses a label, a part of an image, or reference images to guide the model to generate visually plausible images. 
Unlike high-level vision tasks, such as image classification, conditional image generation is a highly subjective task. That is, different generation results are allowed within the scope of visual plausibility.
Such task can be classified into three categories based on the degree of constraints on generated content:
\begin{itemize}
	\item Ill-posed task: There are no constraints on the structure or texture of the generated content, and the only requirement is that the generated samples are visually reasonable. 
	For instance, the inpainting results of a masked face can be either happy or angry.
	\item Semi-ill-posed task: These tasks constrain the structure or texture of the generated result, beyond just being visually reasonable. 
	For instance, given a content image and a style reference, the stylized result is required to be similar to the content image in structure while preserving the stroke/color of the style image.
	\item Well-posed task: These tasks require the generated results to be consistent with the input condition in both structure and texture, and thus relatively fixed generation result is expected. 
	These tasks include small-scale super-resolution, dehazing, denoising, etc.
\end{itemize}
Note that the tasks may not strictly fall into one of the categories. 
For example, in the case of image inpainting with an extremely small masked area 
(e.g., a few pixels), it can be considered as having only one reasonable generation result, and therefore is closer to a well-posed task in the definition.

Traditional conditional image generative methods generate images based on the statistical information of the condition. For instance, super-resolution via bicubic interpolation and face generation based on eigen pattern~\cite{eigenface_turk1991eigenfaces}.
However, for tasks like label-guided image generation, traditional methods often struggle to generate visually plausible results.
In recent years, the development and in-depth study of Generative Adversarial Networks (GANs)~\cite{biggan_brock2018large,stylegan_karras2019style} have been able to exhibit superior performance in generative tasks, even surpassing human-level abilities.
Vanilla GAN samples random noise from a Gaussian distribution $z\sim P_z$ and maps it to the real image distribution $y=f_\theta \left(z\right) \sim P_y$.
Thousands of GAN-based models have been proposed and trained in various subtasks of conditional image generation (e.g., inpainting, style transfer, super-resolution, dehazing) these years.
However, most models designed for these subtasks are deterministic, as the models' inputs are only user-defined deterministic conditions (such as images) rather than randomly sampled noise. 
This implies that for a fixed condition input $x$, the model can only produce a unique corresponding output $y=f_\theta\left(x\right)$, which contradicts the high subjectivity nature of the generation task, especially for the (semi-)ill-posed ones.
Therefore, enabling models of these conditional generative tasks to produce diverse results, and furthermore, controlling the model to produce diverse results satisfying specific requirements (e.g., text guidance), has become a hot research topic.

As summarized in Fig.~\ref{fig:diff_diversity}, there are three mainstream methods for introducing diversity into GAN-based conditional image generative models, which can be divided into two major categories. The first category involves redesigning the non-deterministic model and training, including a) injecting random noise to the model by modulation modules while using loss functions to constrain the one-to-one correspondence between the noise and the generated result~\cite{comodgan_zhao2021large,pic_zheng2019pluralistic} and b) sequentially generating the feature using Gibbs sampling~\cite{put_liu2022reduce,taming_esser2021taming}.
The second category directly employs pre-trained deterministic generative models, as in c) coupling random noise into latent code via well-designed task-related transformation funcations~\cite{wang2021divswapper,cheng2023user}.
Unfortunately, the methods in the first category require users to design modulation modules for different tasks and networks, often demanding substantial computational costs for training.
The method in the second category requires users to design intricate and complex transformation functions. Moreover, since this process can disrupt the inherent structural information of latent codes, it is typically limited to tasks like style transfer where precise reconstruction isn't essential.
The aforementioned diversity methods have strong limitations for real-world applications and often struggle to generate high-quality results. 
\emph{\emph{Recognizing that many existing deterministic generative models have been able to produce satisfactory results, we propose an intriguing question:} for any pre-trained conditional generative model that can only produce high-quality but deterministic result, is it possible to achieve diversity without adjusting the network structure or fine-tuning the parameters?}

The answer is \textbf{yes}. 
Recall the classic task of adversarial attack in the field of AI security, where researchers have found that high-level vision (i.e., image classification, segmentation) models are vulnerable to adversarial examples -- inputs that are almost indistinguishable from natural data and yet classified incorrectly by the network~\cite{fsgm_goodfellow2014explaining}.
For a pre-trained classification network, the attacker only introduces imperceptible perturbations to the input, causing a significant change in the label output. 
Motivated by this, we introduce adversarial attack to low-level vision tasks and find that although the deterministic generative model is robust to random perturbations applied to the input, adversarial examples can still encourage it to generate diverse and visually plausible results, as Fig.~\ref{fig:sample_head} shown.
Therefore, we propose two attacking approaches (i.e., \textbf{untargeted attack} for diverse generation and \textbf{targeted attack} for controllable text/image-guided generation) to empower the existing deterministic generative models to generate diverse and controllable results.

We conducted experiments on different tasks, such as image inpainting, style transfer, and super-resolution, using pre-trained models which have no diversity ability.
Remarkably, without any fine-tuning of the pre-trained generative model, we achieved diverse results and demonstrated an intuitive positive correlation between the model's ability to generate diversity and the non-deterministic nature of the corresponding task.
Additionally, we explored the capability of generating diverse samples with controllable semantics, where we utilized a pre-trained CLIP model~\cite{clip_radford2021learning} to determine the attack direction.
Our contribution can be summarized as:

%-----------------------sample-----------------------
\begin{figure*}[t]
	\centering
	\includegraphics[width=0.9\textwidth]{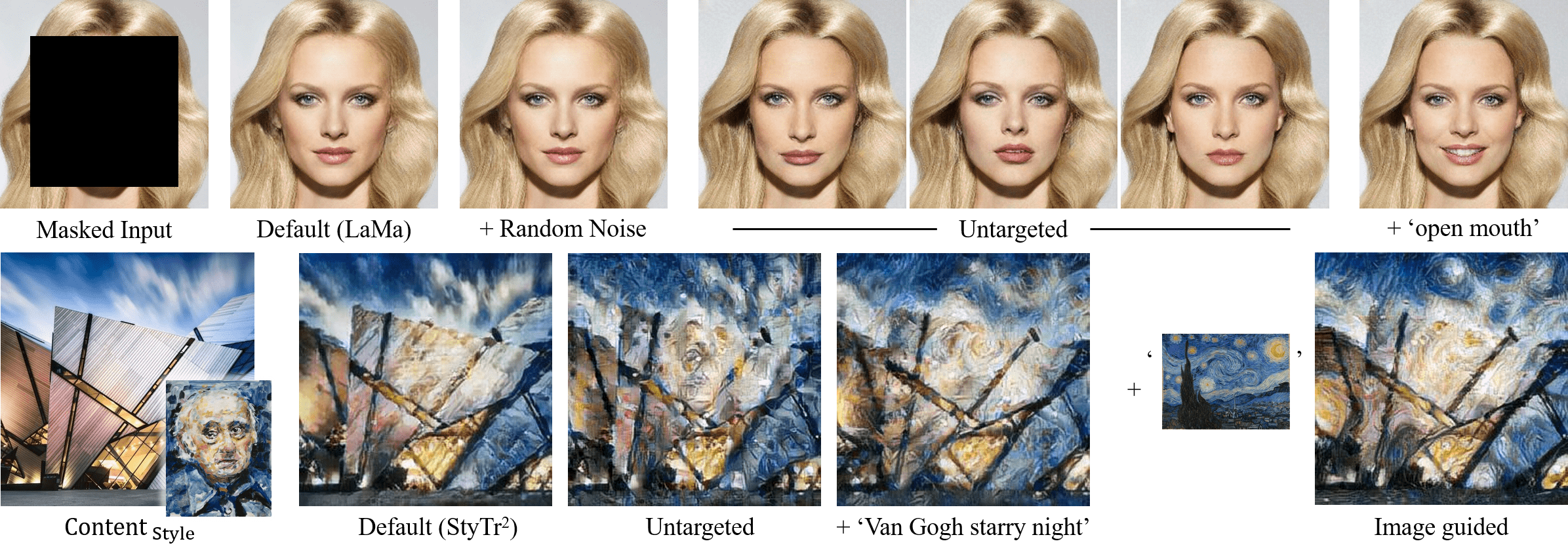}
	\vspace{-0mm}
	\caption{Diverse results generated by our method in conditional image generation tasks.			 
	We have tested upon two pre-trained deterministic models, including LaMa for image inpainting and StyTr$^{2}$ for style transfer. 
	\emph{Random noise \emph{refers to adding standard Gaussian noise to the input.} Untargeted \emph{refers to defining the attack direction to be as different from the default generated results as possible.} +`' \emph{refers to specifying attack direction via text or reference image.}}
		%Our method is capable of inducing diversity in generated outputs, even guiding the model to generate samples that align with the provided descriptions. 
		(zoom-in for details)}
	\label{fig:sample_head}
	\vspace{-0mm}
\end{figure*}
%--------------------------------------------------------------

\begin{itemize}
	\item We first introduce adversarial attack into conditional image generation and demonstrate the potential of deterministic generative models to produce diverse results.
	\item We propose a novel non-learning method that enables diverse generation of a deterministic generative model without any adjustment of network structure or fine-tuning. Our method is plug-in and can be easily applied.
	\item Extensive experiments demonstrate the effectiveness of our method. Our method can guide deterministic/diverse generative model to generate diverse and controllable results thanks to the advantages of being sensitive to initial perturbation and less prone to overfitting the constraint.
	\item Our method provides a new perspective for the interpretability research of low-level vision tasks and vision-language representation model.
\end{itemize}

\section{Related Works}
\subsection{Conditional Image Generation}
Traditional conditional image generation algorithms~\cite{hertzmann2003survey} aim to solve well-posed problems, e.g., image completion of simple geometric structure.
With the development of deep learning,~\cite{first_style_johnson2016perceptual, xie2012image} took the lead in applying deep neural networks to conditional image generation and achieved impressive performance.
\cite{cgan_mirza2014conditional} first introduced generative adversarial network (GAN) structure to conditional image generative model, expanding new ideas for subsequent researchers.
\cite{isola2017image} exploits conditional GANs for inpainting. 
\cite{edgeconnect_nazeri2019edgeconnect, yang2020learning} proposed using coarse (edge map, gradient map, etc.) generation results to guide the inpainting.
Additionally,~\cite{gateconv_yu2019free, LaMa_suvorov2022resolution} have explored the impact of different computational modules on inpainting performance.
\cite{gatys2016image} discovered that the Gram matrix upon deep
features extracted from a pre-trained DCNN can notably represent the characteristics of visual styles, which opens up the era of neural style transfer.
\cite{first_gan_styleTransfer_sanakoyeu2018style} introduced GAN structure for style transfer. 
Subsequent works improve the performance of neural style transfer in many aspects, including quality~\cite{stytr2_deng2021stytr} and generalization~\cite{chiu2019understanding}.
\cite{dong2015compression} took the lead in introducing learning-based method into well-posed vision tasks, e.g., super-resolution, denoising, and JPEG compression artifact reduction.
Subsequent works~\cite{psd_chen2021psd,swinir_liang2021swinir} have explored the impact of network structure on these tasks.

However, unlike probabilistic models, these methods cannot rely on random sampled input, which limits their ability to generate diverse samples for a fixed condition.
The lacking of diversity can sometimes be unacceptable, especially in real-world applications of ill-posed tasks,  even if those methods can produce high-quality results.
Our method enables deterministic conditional generative models to produce diverse outputs with simple operations and minimal computational cost via adversarial attack and even guides them to generate outputs with specified prompts or reference, even if the conditions were never labeled in the training set.

\subsection{Diverse Image Generation}
Vanilla GAN~\cite{gan_goodfellow2020generative,stylegan_karras2019style} belongs to probabilistic model, which learned to map from a normal distribution to the complex distribution of real image. 
Unlike vanilla GANs, conditional image generative models learn to map conditions to real images, where the conditional is deterministic and cannot be obtained through random sampling, which makes it difficult to produce diverse outputs. 
Previous researchers have proposed three mainstream methods for introducing diversity into conditional image generative models:
1) Injecting noise into the bottleneck layer directly or via a rather small modulation network and constraining the noise-output relationship through loss functions~\cite{comodgan_zhao2021large,pic_zheng2019pluralistic}. 
Such methods generate diversity through randomly sample the noise during inference, which require training the network along with the noise or fine-tuning a pre-trained generator.
2) Sequentially generating a sequence using Gibbs sampling, which is commonly used in transformer models~\cite{put_liu2022reduce,taming_esser2021taming}. The next predicted token's probability distribution is determined by the previously generated tokens due to the Markov chain property, leading to diversity. 
This method is usually time-consuming due to the inability to parallelize sampling.
3) Applying feature transformation in the latent space~\cite{wang2021divswapper,cheng2023user}. This method requires careful designing of transformation functions that may significantly affect the network's ability to reconstruct texture and structure accurately.
As such, it is only applicable in style transfer tasks with a high tolerance for the lacking of reconstruction ability.

The above methods require high computational costs and usually can only produce diversity on a specific subtask or a specific model.
Our method is applicable to most of the conditional image generative models (especially GAN-based) without the need of re-training, fine-tuning, or designing any complex transformation function.

\subsection{Adversarial Attack}
The concept of adversarial attack is first proposed by~\cite{first_adv_szegedy2013intriguing}, which are constructed by adding perturbations that are too small to be recognized by human eyes to an image but could cause the misclassification of the classification model with high confidence.
Adversarial attack is typically applied in image classification~\cite{papernot2016limitations}, object detection~\cite{xiao2018characterizing}, machine translation~\cite{belinkov2017synthetic}, etc.

Works such as~\cite{fsgm_goodfellow2014explaining,pgd_madry2017towards} explore white-box attack methods, which allow attackers to access model parameters and gradients. 
In contrast,~\cite{hsja_chen2020hopskipjumpattack} explore black-box attacks, which only provide attackers with access to the model's input and output.
These works are usually used to explore the linearity, robustness, and interpretability of deep neural networks. Our method is the first to introduce adversarial attack into conditional image generation task, bringing diversity to deterministic generative models in a simple but effective way.

%%-----------------------advattack-----------------------
%\begin{figure}[t]
%	\centering
%	\includegraphics[width=0.99\linewidth]{advattack}
%	\vspace{-0mm}
%	\caption{Scheme of targeted multi-step adversarial attack.
%		Pre-trained CLIP model is utilized to determine the direction of adversarial attack.
%	}
%	\label{fig:adv}
%	\vspace{-1mm}
%\end{figure}
%%--------------------------------------------------------------

\section{Method}
For a deterministic conditional image generative network $f$ with parameter weight of $\theta$ and input condition $x$, the goal of adversarial attack is to introduce smallest possible perturbation $z$ to input $x$ to make the result $y'=f_{\theta}\left( x+z\right)$ as different as possible from the default output $y=f_{\theta}\left( x\right)$.

%\subsection{White-box attack}
%When applying white-box attack, the attacker is allowed to access all information about the target model, including network structure and parameters.
%Since gradient can be obtained from the above information, attack direction can be rather easily determined.
%In this paper, we mainly focus on discussing this form of attack, which can be categorized into untargeted attack and targeted attack.
\subsection{Untargeted Attack for Diversity}
To perform adversarial attack on conditional image generative models, we need to obtain specific perturbations tailored to the input condition $x$.
Following fast gradient sign method (FGSM)~\cite{fsgm_goodfellow2014explaining}, which proposed to linearize the loss function $\mathcal{L}$ around the weight $\theta$ of the model, obtaining an optimal max-norm constrained perturbation $z=\epsilon sign\left(\nabla_x \mathcal{L}\left( \theta,x,y\right)\right)$, we define $x+z$ as our adversarial sample.
Unlike image classification, it is challenging to design an attack loss that aligns with the training loss (e.g., GAN loss) of the image generative model. 
However, we have found that using simple statistical features as the attack loss can already generate samples that differ from the default results. 
Two simple statistical losses are used in our paper for untargeted attack.
\begin{equation}
	\begin{aligned}
		\mathcal{L}_{L1}\left( \theta,x,y\right)=\|f_\theta\left(x\right)-y\|_1\\
		\mathcal{L}_{var}\left( \theta,x,y\right)=var\left(f_\theta\left(x\right)\right)
	\end{aligned}
	\label{con:L1}
\end{equation}
$y$ is the default generated result, $var$ indicates deviation var of a generated image pixels. Gradient $\nabla_x$ can be easily obtained via backpropagation. $\mathcal{L}_{L1}$ is used as the statistical feature based losses in the following formulas.

The above process can be extended into multi-step for better attack performance as suggested in~\cite{pgd_madry2017towards}.
We initialize the attack direction~\cite{wong2020fast} by adding a micro Gaussian noise $z_0$ on the original condition $x$. 
The adversarial sample is truncated by $c_{min}$ and $c_{max}$ in case of exceeding the value range.
The process of multi-step adversarial attack can be expressed as:
\begin{equation}
	\begin{aligned}
		&x_1=x+z_0,\ \ z_0\sim N(0,\delta^2)\\
		&z_i=\nabla_{x_i}\mathcal{L}_{L1}(\theta,x_i,y)\\
		&x_{i+1}=clip\left(x_i+\epsilon sign\left(z_i\right),c_{min},c_{max}\right)
	\end{aligned}
\end{equation}
However, we found that the attack may lead to serious artifacts in the generated results since $\mathcal{L}_{L1}$ not considering the visual plausibility of generated result. Therefore, we propose to use noise truncation instead of the PGD truncation method (see supplementary material for visual comparison):
\begin{equation}
	\begin{aligned}
		x_{i+1}=x+clip\left(x_i-x+\epsilon sign\left(z_i\right),c_{min},c_{max}\right)
	\end{aligned}
\end{equation}

\begin{algorithm}[tb]
	\caption{Adversarial attack on deterministic generative model, given pre-trained generative model $f_\theta$ and cross-model Vision-language representation model $\mathit{CLIP}$.
		\label{alg:main}}
	\label{alg:algorithm}
	\textbf{Input}: Original input condition $X=\left\{x^1,\dots,x^n\right\}$\\
	\textbf{Output}: Generation result $Y$.
	\begin{algorithmic}[1] %[1] enables line numbers
		\STATE Random initialize each condition $x^k_1=x^k+z^k_0,$\\$ z^k_0\sim N(0,{\delta^k}^2),\ \ k=1,\dots,n$.\\
		\STATE Init $\left\{\epsilon^1,\dots,\epsilon^n\right\},\ c_{min},\ c_{max},\ {i,j} = 1$ \\
		\WHILE{$i<step$}
		\WHILE{$j<n$}
		%\STATE Do some action.
		\IF {untargeted}
		\STATE $z^j_i=\nabla_{x^j_i}\mathcal{L}_{L1}\left(\theta,X_i,y\right) $\ \ // Equation~\eqref{con:L1}
		\ELSE
		\STATE {$z^j_i=\nabla_{x^j_i}\mathcal{L}_{\mathit{CLIP}}\left(\theta,X_i,y\right) $\ \ // Equation~\eqref{con:sim}}
		\ENDIF\\
		\STATE $\delta_x = \mathit{clip}\left(x^j_i-x^j+\epsilon^j\mathit{sign}\left(z^j_i\right),c_{min},c_{max}\right)$\\
		\STATE $x^j_{i+1}=x^j+\delta_x$\\
		\STATE $\epsilon^j=\epsilon^j * 0.95$\\
		\STATE $j = j + 1$\\
		\ENDWHILE\\
		\STATE $Y_{i+1}=f_\theta\left(X_{i+1}\right)$\\
		\STATE $i = i + 1$\\
		\ENDWHILE
		\STATE \textbf{return} $Y_{i}$
	\end{algorithmic}
\end{algorithm}
\subsection{Targeted Attack for Controllability}
There has been a significant amount of work exploring multimodal image synthesis/editing~\cite{zhan2021multimodal}, among which using image or text to guide image generation is one of the most commonly used methods.
Recently proposed Contrastive Language-Image Pre-training (CLIP)~\cite{clip_radford2021learning} model has shown remarkable ability in extracting semantic correspondences between image and text.
Hence we leverage its capabilities in our method to achieve targeted attack for image/text-guided generation.
%-----------------------face-----------------------
\begin{figure*}[t]
	\centering
	\includegraphics[width=0.97\textwidth]{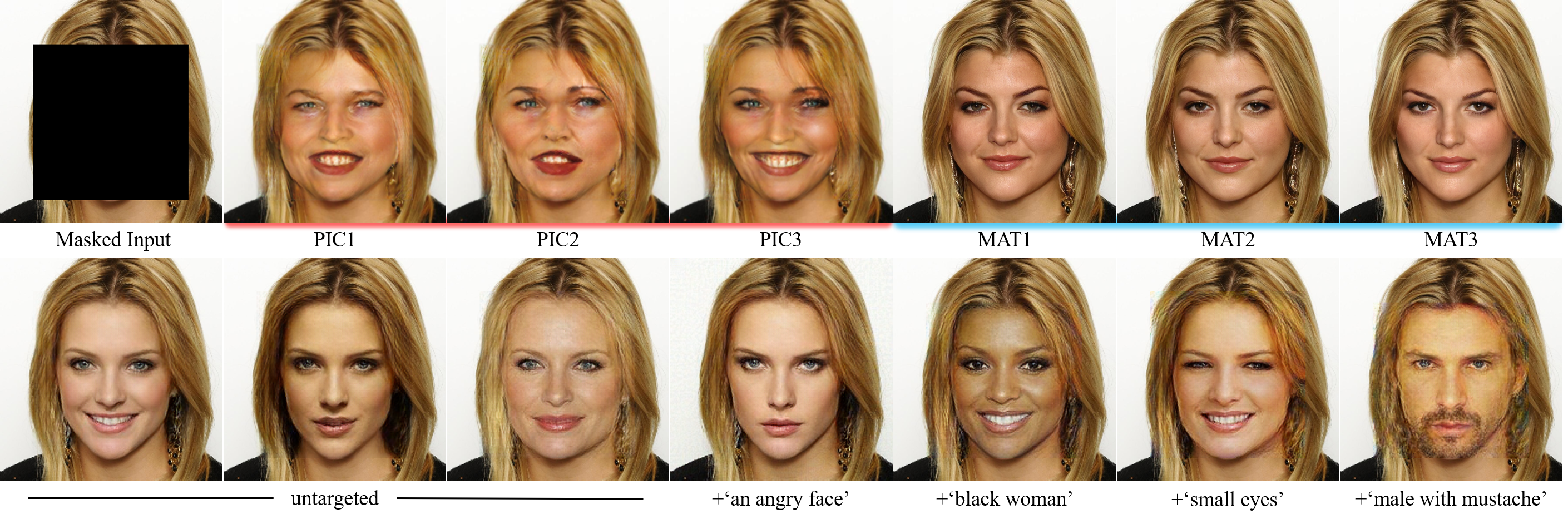}
	\vspace{-3mm}
	\caption{Diverse face inpainting results generated by attacking the deterministic inpainting model LaMa. 
		Generated results are compared with diverse inpainting model PIC and MAT.
	}
	\label{fig:inpainting}
	\vspace{-0mm}
\end{figure*}
%--------------------------------------------------------------
We align the vector pointing from the default generated result to the attacked generated result with the vector from the source image's description to the target description in the CLIP space, since directional CLIP loss~\cite{stylegan_nada_gal2022stylegan} is proved to have better guidance ability compared to general CLIP similarity loss.
Targeted attack direction can be defined as:
\begin{equation}
	\begin{aligned}
		\triangle I=\mathit{CLIP}_I\left(f_\theta\left(x_i\right)\right) - \mathit{CLIP}_I\left(y\right)
	\end{aligned}
\end{equation}
\begin{equation}
	\begin{aligned}
		\triangle R=\left \{
		\begin{array}{ll}
			\mathit{CLIP}_T\left(T_{ref}\right) - \mathit{CLIP}_T\left(T_{src}\right),                    & text\\
			\mathit{CLIP}_I\left(I_{ref}\right) - \mathit{CLIP}_I\left(y\right),     & image\\
		\end{array}
		\right.
	\end{aligned}
\end{equation}
\begin{equation}
	\begin{aligned}
		\mathcal{L}_{\mathit{CLIP}}\left(\theta,x_i,y\right)=\frac{\triangle I \cdot \triangle R}{|\triangle I||\triangle R|}
	\end{aligned}
	\label{con:sim}
\end{equation}
In which, $\mathit{CLIP}_I$ and $\mathit{CLIP}_T$ are CLIP's image and text encoders, $T_{ref}$ and $T_{src}$ are target and source text description, $I_{ref}$ is the reference image.
When image is used as reference, directional CLIP loss degrades to the general clip similarity loss.
Pseudocode of untargeted/targeted attack can be seen in Alg.~\ref{alg:main}.

\section{Experiments}
We demonstrated the effectiveness of our method on different types of conditional image generation tasks.
SOTA pre-trained deterministic models, including LaMa~\cite{LaMa_suvorov2022resolution} for inpainting and StyTr$^2$~\cite{stytr2_deng2021stytr} for style transfer, are used to qualitatively and quantitatively validate the feasibility and effectiveness of our proposed method. 
In addition, we also conducted experiments on tasks such as super-resolution, dehazing, and probabilistic generation to further discuss the generalizability and limitations of our proposed method.
The attack step is set to 10 by default.

%-----------------------SR-----------------------
\begin{figure}[t]
	\centering
	\includegraphics[width=0.47\textwidth]{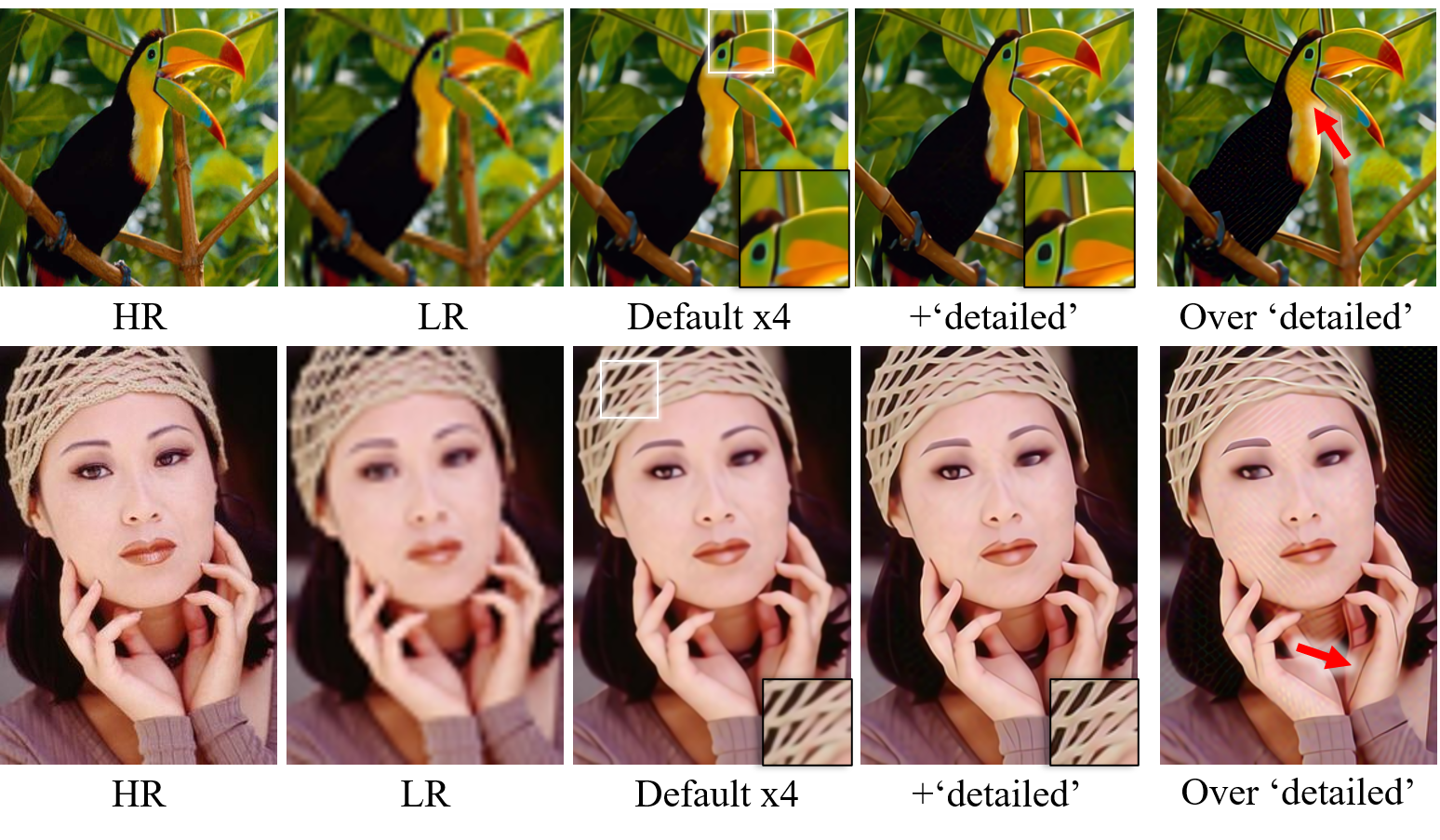}
	\vspace{-5mm}
	\caption{Our method works well on super-resolution (well-posed vision task). 
		SwinIR x4 model demonstrates sharper generated results via attack using "detailed" as the direction.
		%However, for dehazing model PSD~\cite{psd_chen2021psd}, adversarial samples tend to introduce artifacts rather than producing satisfactory guided generative results. 
	}
	\label{fig:SR}
	\vspace{-0mm}
\end{figure}
%--------------------------------------------------------------

%-----------------------style-----------------------
\begin{figure*}[t]
	\centering
	\includegraphics[width=0.97\textwidth]{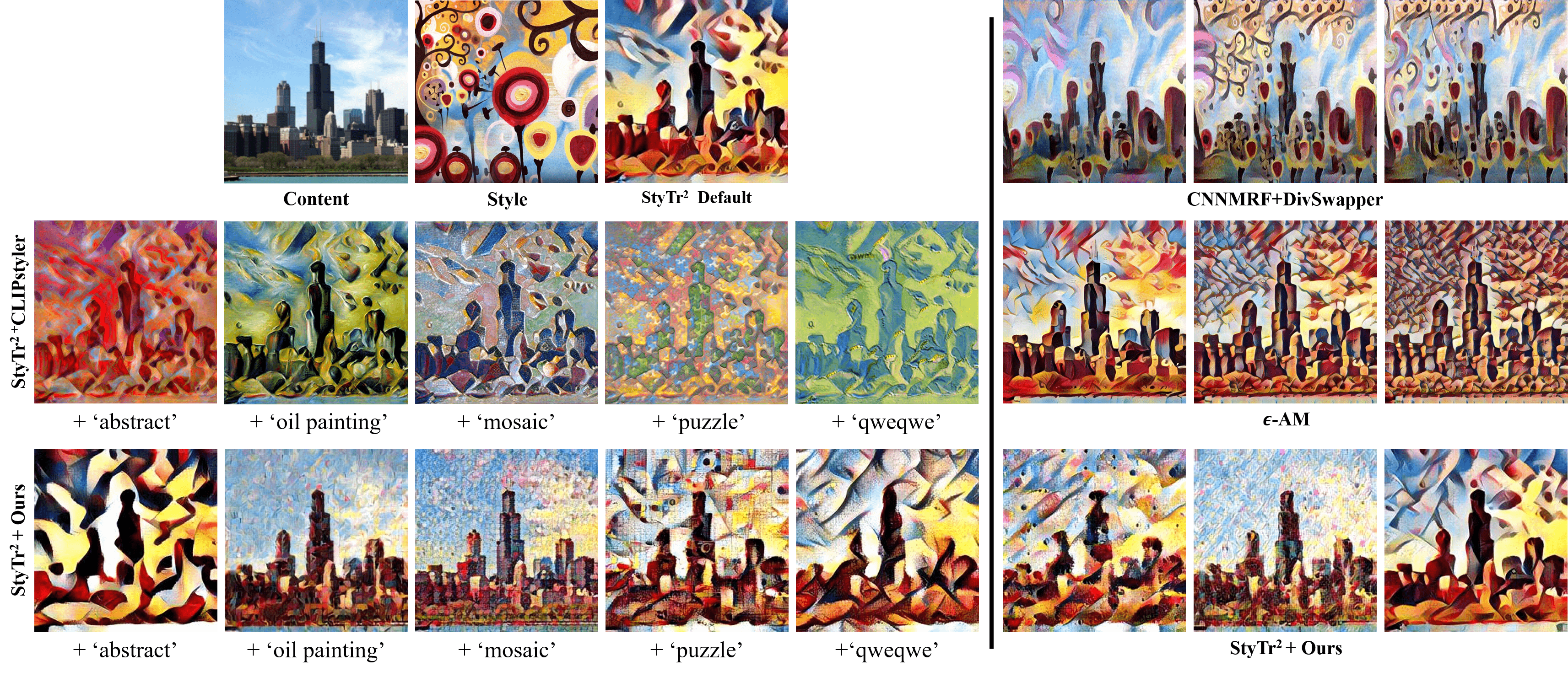}
	\vspace{-0mm}
	\caption{Left: Targeted diverse stylization.
		Top row: Content image, style image, and the default stylized result of StyTr$^2$.
		Second and third row: Text-guided stylization, compared with CLIPstyler~\cite{clipstyler_kwon2022clipstyler} which also uses CLIP for guidance. The default stylized image in row one is used as the input of CLIPstyler. Our method faithfully preserves the color characteristics of the style image. 
		CLIPStyler requires fine-tuning the reconstruction model, which takes several minutes, while our method can complete each step of attack within 0.2 second.
		Right: Untargeted diverse stylization. Compared with DivSwapper and $\epsilon$-AE, our method generates higher diversity with better quality.
	}
	\label{fig:style}
	\vspace{-0mm}
\end{figure*}
%--------------------------------------------------------------

\begin{table}[t]
	\vspace{-3mm}
	\label{tab:diversity}
	\centering
	\renewcommand{\arraystretch}{1.29}
	\resizebox{\linewidth}{!}{
		\begin{tabular}{lcccc}
			\hline
			inpainting  & LPIPS$\uparrow$ & L1$\uparrow$    & FID   & Runtime        \\ \hline
			LaMa~\cite{LaMa_suvorov2022resolution}                            & 0               & 0               & 50.37 & 0.19s          \\
			LaMa+Ours                       & 0.0606          & 13.5963         & 52.98 & 1.99s(10 step) \\
			MAT~\cite{mat_li2022mat}                             & 0.0176(29.0\%)  & 5.9999(44.1\%)  & 45.93 & 0.23s          \\
			\begin{tabular}[c]{@{}l@{}}StableDiffusion-v1.5\\ \cite{sd_rombach2022high}\end{tabular}           & 0.0634(104.6\%) & 12.4692(91.7\%)   & 146.60 & 5.5s           \\
			ICT~\cite{ict_wan2021high}               & 0.0498(82.2\%)  & 12.5915(92.6\%)   & 75.41 & 89s            \\
			BAT~\cite{bat_yu2021diverse}                             & 0.0390(64.4\%)  & 10.7356(78.96\%)  & 62.44 & 22s            \\
			PIC~\cite{pic_zheng2019pluralistic}                             & 0.0265(43.7\%)  & 8.4986(62.5\%)  & 72.64 & 0.19s          \\ \hline
			style transfer                  &                 &                 &       &                \\\hline
			StyTr$^2$~\cite{stytr2_deng2021stytr}                       & 0               & 0               & -     & 0.10s          \\
			StyTr$^2$+Ours                  & 0.3199          & 51.6725         & -     & 1.10s(10 step) \\
			$\epsilon$-AE~\cite{cheng2023user}                  & 0.1826(57.1\%)  & 37.1325(71.9\%) & -     & 1.10s          \\
			%CNNMRF+DivSwapper~\cite{wang2021divswapper}               & 0.2183(68.2\%)  & 39.1017(75.7\%) & -     & 23.4s          \\
			\begin{tabular}[c]{@{}l@{}}CNNMRF+DivSwapper\\ \cite{wang2021divswapper}\end{tabular}               & 0.2183(68.2\%)  & 39.1017(75.7\%) & -     & 23.4s          \\
			Avatar-net+DFP~\cite{wang2020diversified}					& 0.2044(63.9\%)  & 42.5471(82.3\%)  & -     & 3.5s           \\ \hline
		\end{tabular}
	}
	\caption{Quantitative comparison of inpainting and style tranfer. Higher LPIPS/L1 score means higher diversity. Our method employs the standard 10-step attack, all experiments were conducted on a single RTX 3090 GPU.}
	\vspace{-4mm}
\end{table}

\subsection{Qualitative and Quantitative Comparison}
For image inpainting (ill-posed task), we use $\epsilon\in\left[0.01,0.05\right]$ with $c_{min}=-0.09,\ c_{max}=0.09$. As shown in Fig.~\ref{fig:inpainting}, PIC~\cite{pic_zheng2019pluralistic} can produce relatively higher diversity but lacks the ability to inpaint large masked areas, resulting in poor quality of generated samples. MAT~\cite{mat_li2022mat} generates high-quality samples but can only achieve subtle diversity, which is hard to perceive. Our method achieves a satisfactory balance between generation quality and diversity.
For style transfer (semi-ill-posed task), we use $\epsilon\in\left[0.1,0.25\right]$ with $c_{min}=-0.25,\ c_{max}=0.25$. As shown in Fig.~\ref{fig:style}, diverse samples generated by $\epsilon$-AE~\cite{cheng2023user} share the same stroke patterns and may suffer from severe distortion under inappropriate hyperparameter settings, which fail to reflect the correlation between stylized image and style reference. 
DivSwapper~\cite{wang2021divswapper} generates diversity through latent space swapping, but it is still limited by the pattern of the default stylized image and inevitably produces repeated textures.
Our method maintains color consistency while generating style diversity.
Benefiting from the prior of CLIP, our method can even make the stylized image have features that are not present in the style reference, such as "mosaic" or "puzzle" style in Fig.~\ref{fig:style} (refer to supplementary material for more experimental results).
%Furthermore, by using the style image as the attack direction, we can make the stylized results more stylized, for example, stylized results obtained distorted lines after using the text of 'van gogh starry night'.
To quantify diversity, we report the LPIPS distance and L1 distance between diverse generated samples in Tab.~\ref{tab:diversity}. 
It can be observed that samples generated by our method have larger differences between each other, indicating higher diversity.

A reasonable assumption is that conditional image generative models have the ability to produce diversity if the loss function is loose (e.g., making the generated results conform to a certain distribution), which allows the model enough flexibility, so that our method can make (semi-)ill-posed vision models produce obvious diversity, as shown in Fig.~\ref{fig:inpainting} and~\ref{fig:style}.
Surprisingly, our method also works well for the well-posed vision task models that only limited diversity can be accepted. 
As shown in Fig.~\ref{fig:SR}, when apply `detailed' with $\epsilon=0.005$ and $c_{min}=-0.01,\ c_{max}=0.01$ to the SwinIR~\cite{swinir_liang2021swinir} model, we successfully encourage it to generate sharper results than the default ones. (refer to supplementary material for quantitative comparison)

\subsection{Compared with Optimizer-based Method}
We noticed that early works in style transfer~\cite{gatys2016image} utilized optimizer to add gradient information of style representations from different layers of pre-trained neural networks to the content image for style transfer. In order to further explore the performance differences between our proposed method and the optimizer-based method, we treat the input condition $x$ as the parameter to be optimized and use Adam optimizer for gradient propagation. 
In comparison to the optimizer-based method, our method has the following advantages:

1) Our method produces better diversity, which is attributed to its sensitivity to the initial perturbation. When applying small random perturbations to the input, our method can obtain different paths of variation, resulting in more diverse generated samples.
We tested five sets of perturbation on LaMa with the same input masked images and an initial perturbation of $z_0\sim N\left(0,10^{-14}\right)$ using our method and the optimizer-based method. 
We calculated the L1 distance between each pair of updated perturbation $\|x_{i+1}-x_i\|_1$, where the average distance were $3.9\times10^{-3}$ for our method and $2.54\times10^{-5}$ for optimizer-based method. This confirms the sensitivity of our method to the initial perturbation.

\begin{figure}[t]
	\centering
	\includegraphics[width=0.47\textwidth]{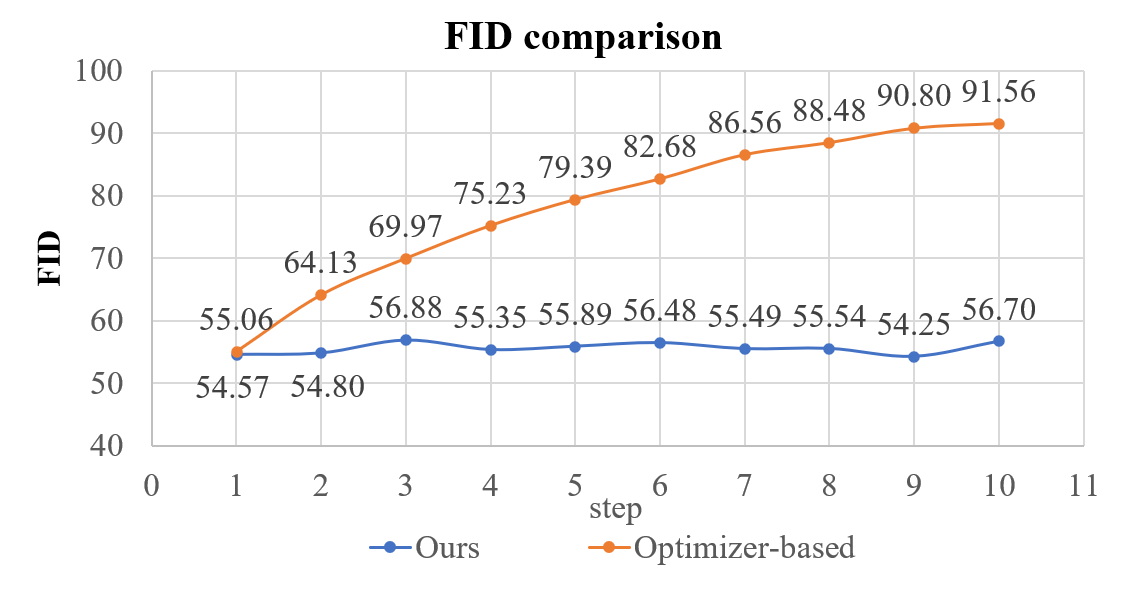}
	\vspace{-5mm}
	\caption{FID comparison of each step between our method and optimizer-based method to generate diverse inpainting result via LaMa.
	}
	\label{fig:FID}
	\vspace{-0mm}
\end{figure}

2) Our method demonstrates better generation quality, which is specifically reflected in 1) less prone to overfit the constraint and 2) lesser prone to artifacts.
Even after a sufficient number of iterations, our method can still generate visually plausible results, whereas the optimizer-based method tends to produce visually implausible ones. 
As shown in Fig.~\ref{fig:FID}, our method maintains a relatively stable FID score as the iteration number increases, while the quality of the optimizer-based method deteriorates significantly.

%-----------------------FID-----------------------
%--------------------------------------------------------------
%-----------------------gans-----------------------
\begin{figure}[t]
	\centering
	\includegraphics[width=0.47\textwidth]{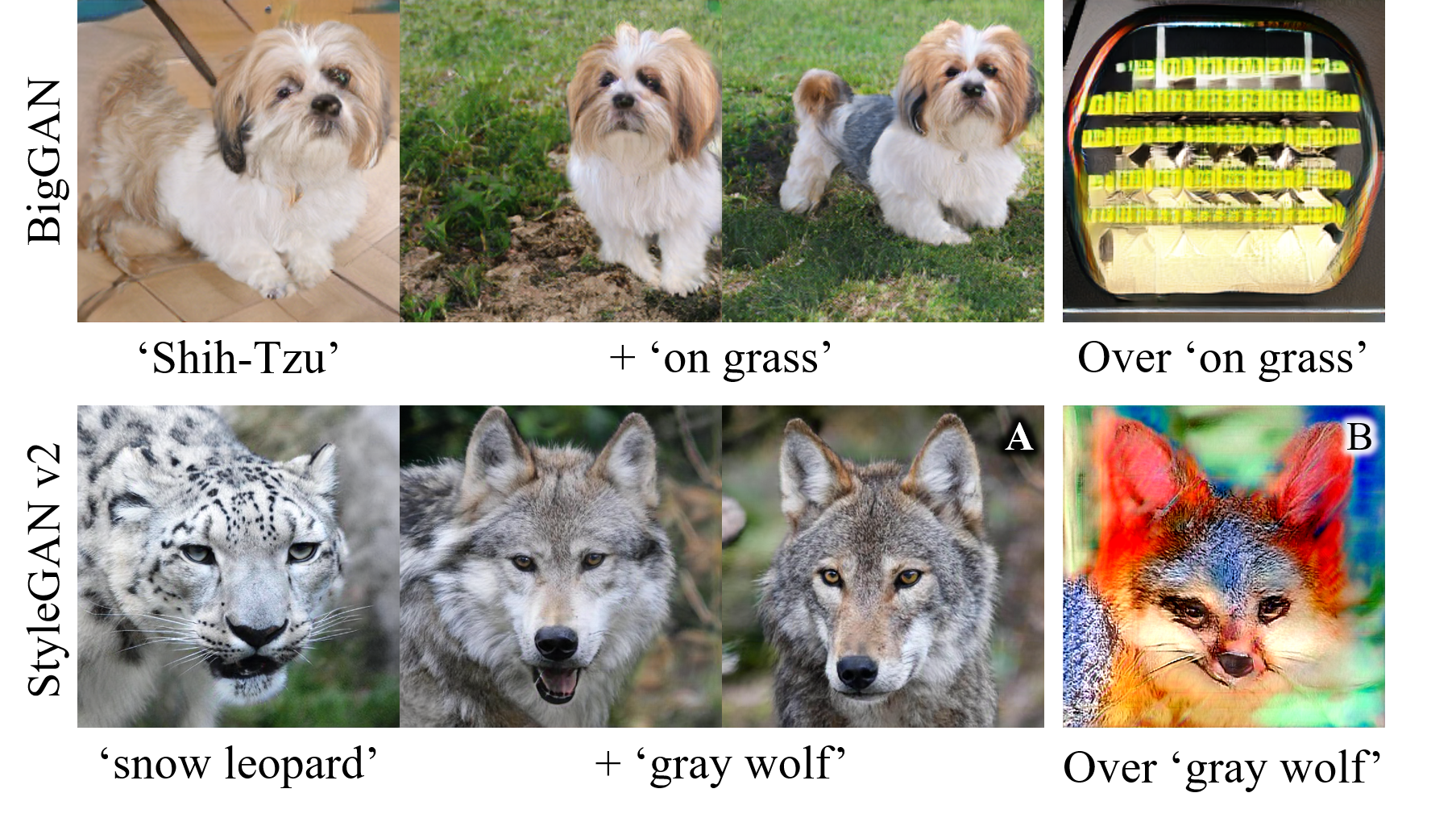}
	\vspace{-6mm}
	\caption{Our method also be applied to semantic control of probabilistic generative models. We observed that the model tends to generate unreasonable results when $\epsilon$ is too large.
	}
	\label{fig:GANs}
	\vspace{-0mm}
\end{figure}
%--------------------------------------------------------------

\subsection{Compared with Diffusion-based Method}
We have noticed the remarkable performance of recently proposed diffusion models. Those models have been trained on extremely large text-image datasets and have the ability of open-domain generation.
In our comparisons with diffusion models, we observed the following:
1) The generative performance of diffusion models heavily depends on the prompt. When using simple prompts like `an angry face' as used in Fig.~\ref{fig:inpainting}, diffusion models tend to generate content that aligns with the prompt but lacks visual coherence.
2) Diffusion models can not be directly applied to various subtasks of conditional image generation. Often, specific feature injection networks like ControlNet are required.
3) Diffusion models are constrained by multi-step reasoning, which often leads to longer inference times to generate high-quality results. (refer to supplementary material for more comparison)

\subsection{Other Discussions}
When generating adversarial examples for generative models, we expect the attack noise to be as small as possible (ideally, imperceptible to the human eyes) and the generated result to be visually plausible. 
Interestingly, when the attack noise is over large but not dominant in the input condition, the non-robustness of the generative model to adversarial examples can lead to the appearance of essential features of the model.
As can be seen in Fig.~\ref{fig:SR}, when a super-resolution model generates overly `detailed' results, it is performing operations such as dividing different colors apart, enhancing or adding boundaries and adding grid-like artifacts.
It can also be observed in the failure cases (refer to supplementary material) of inpainting model that the model tends to select one of the learned ``standard pattern" and splice it onto the degraded area, similar to eigenface~\cite{eigenface_turk1991eigenfaces} in some extend. 
This also provides a novel view for the interpretability and data security study of conditional generative models.

We also conducted experiments on BigGAN~\cite{biggan_brock2018large} and found that no matter how much random noise was added to the input, the model always generated samples that matched the class label. 
However, when we used adversarial examples with larger $\epsilon$ values (while keeping the distribution $x+z_i$ same to $x$), the model generated meaningless samples that did not match the label. This demonstrates that the mapping from the latent space to the image space that BigGAN has learned is incomplete. (see the detailed experiment in supplementary material)

Another interesting phenomenon we observed is that the CLIP space and human perceptual space are not strictly aligned. When the attack strength is set excessively high while not constraining the overall perturbation, the output tends to overfit the text guidance. For example, in Fig.~\ref{fig:GANs}, the CLIP similarity of sample A with the text "gray wolf" is 0.2030, while that of B is 0.2185. which means in CLIP space, sample B is more `gray wolf' than sample A.
We believe this is also the reason why our method outperforms the optimizer-based method for attacking deterministic models to generate diverse results.

%%-----------------------biggan-----------------------
%\begin{figure*}[t]
%	\centering
%	\includegraphics[width=0.94\textwidth]{biggan}
%	\caption{Our method can be applied on diverse image synthesis model like BigGAN~\cite{biggan_brock2018large}.
%	}
%	\vspace{-5mm}
%	\label{fig:biggan}
%	\vspace{1mm}
%\end{figure*}
%%--------------------------------------------------------------

%
%%-----------------------interpretability-----------------------
%\begin{figure*}[t]
%	\centering
%	\includegraphics[width=\textwidth]{inter}
%	\caption{Failure cases. When applying inappropriate attack settings, the generated results exhibit special patterns, which can help us understand what the generative model has learned.
%	}
%	\label{fig:inter}
%	\vspace{-2mm}
%\end{figure*}
%%--------------------------------------------------------------

\section{Conclusion}
We propose a simple and efficient approach to generating adversarial examples for generative models without requiring any modifications to their parameters or architectures. 
We first utilized adversarial attack to induce diversity in existing pre-trained deterministic conditional image generative models. 
Additionally, we leverage a pre-trained CLIP model to control the attack direction and encourage the generation of samples that satisfy specific semantics. 
We evaluate our method on various generative tasks and demonstrate that it achieves results surpassing those of state-of-the-art diverse generative models. 
Finally, we discuss the potential of adversarial attack in the interpretability and data security of low-level vision models.

\section{Acknowledgments}
This work was supported in part by Zhejiang Province Program (2022C01222, 2023C03199, 2023C03201, 2019007, 2021009), the National Program of China (62172365, 2021YFF0900604, 19ZDA197), Ningbo Program(2022Z167), and MOE Frontier Science Center for Brain Science \& Brain-Machine Integration (Zhejiang University).

\bibliography{aaai24}

\end{document}